\documentclass{llncs}
\usepackage[T1]{fontenc}
\usepackage{graphicx}

\usepackage{booktabs}
\usepackage{multirow}
\usepackage{amsfonts}
\usepackage{xcolor}
\usepackage{amsmath}
\usepackage[hidelinks]{hyperref}
\usepackage{footmisc}

\usepackage{color}
\begin{document}
\title{Mapping the Media Landscape: Predicting Factual Reporting and Political Bias Through Web Interactions}

%\titlerunning{Predicting Factual Reporting and Political Bias Through Web Interactions}
% If the paper title is too long for the running head, you can set
% an abbreviated paper title here
%

%\author{Anonymous CLEF submission}
%\orcidID{0000-0003-2429-6152}

\author{Dairazalia Sánchez-Cortés\inst{1} \and
Sergio Burdisso\inst{1} 
\and
Esaú Villatoro-Tello\inst{1}
\and
Petr Motlicek\inst{1,2}
}

% \authorrunning{F. Author \textit{et al.}}

% First names are abbreviated in the running head.
% If there are more than two authors, '\textit{et al.}' is used.

\institute{Idiap Research Institute, Martigny, Switzerland \and
Brno University of Technology, Brno, Czech Republic\\
\email{\{dscortes,sergio.burdisso,esau.villatoro,petr.motlicek\}@idiap.ch}}

\maketitle              % typeset the header of the contribution
\begin{abstract}

Bias assessment of news sources is paramount for professionals, organizations, and researchers who rely on truthful evidence for information gathering and reporting. While certain bias indicators are discernible from content analysis, descriptors like political bias and fake news pose greater challenges. In this paper, we propose an extension to a recently presented news media reliability estimation method that focuses on modeling outlets and their longitudinal web interactions.
Concretely, we assess the classification performance of four reinforcement learning strategies on a large news media hyperlink graph.
Our experiments, targeting two challenging bias descriptors, factual reporting and political bias, showed a significant performance improvement at the source media level. 
Additionally, we validate our methods on the CLEF 2023 \textit{CheckThat!} Lab challenge, outperforming the reported results in both, F1-score and the official MAE metric. 
Furthermore, we contribute by releasing the largest annotated dataset of news source media, categorized with factual reporting and political bias labels. Our findings suggest that profiling news media sources based on their hyperlink interactions over time is feasible, offering a bird's-eye view of evolving media landscapes.\footnote{\href{https://github.com/idiap/Factual-Reporting-and-Political-Bias-Web-Interactions}{github.com/idiap/Factual-Reporting-and-Political-Bias-Web-Interactions}}

\end{abstract}
%available Topics: 
%multilingual multimodal information access 
%--->evaluation methodologies 
%analytics for information retrieval 
%reproducibilty and replicability issues
%
%
%
\section{Introduction}
Given its open and distributed nature, the World Wide Web (WWW) has become the main information source worldwide, democratizing content creation and making it easy for everybody to share and spread information online. On the bright side, this phenomenon enables a faster dissemination of information compared to what was possible with traditional newspapers, radio, and TV. On the downside, at the moment of removing the "gate-keeper" role from traditional media, it opens the door for additional problems, e.g., the spread of misinformation, at breaking-news speed, that can potentially mislead the users and even impact their behavior \cite{stromback2020news,baly-etal-2018-predicting}. 

Thus, while the goal of this democratic channel is to provide users with the necessary tools to acquire greater knowledge about a topic, the reality is that in the way this knowledge (i.e., news) is presented and reported is not necessarily always impartial \cite{azizov2023frank,kulshrestha2019search}, and there is a growing concern regarding the biases of different media outlets when reporting specific events \cite{Hamborg2023}. For example, in polarizing topics like politics, many of the news can be biased towards one political perspective or the other, i.e.,  \textit{political bias}, which may influence citizens’ voting decisions and preferences of undecided individuals \cite{gezici2022quantifying}.

To mitigate the impact of misinformation and to favor critical assessment for the newsreaders, independent bias assessment services like  MBFC\footnote{\label{note1} \url{https://mediabiasfactcheck.com}} and allsides\footnote{\label{noteallsides}\url{www.allsides.com}} perform information verification. The review process is performed manually by professionals at the event or article level, clearly this is a challenging schema to maintain on the long term given the fast-speed proliferation of both news media websites and news articles. 
Automation comes handy to perform certain fact-checking tasks, like gathering information (e.g. articles with similar topics, metadata on the media-publisher, etc.); for the more complex parts of the verification analysis, advances in AI continues pushing the boundaries in order to provide valuable tools (for example, search and retrieval, summarization, transformers, LM and LLMs). 
While the latest LLMs performance on several tasks is remarkable, they are still prone to carry unauthenticated information~\cite{kamalloo2023evaluating}.

While many existing tools are being adopted to support verification tasks at the article level (with and without human supervision), there are very few advances to fully automate news media profiling at the source level (other than popularity). 
Previous research has shown evidence that some news bias descriptors can be inferred by just inspecting the outlet website metadata \cite{esteves2018belittling,hounsel2020identifying}. Other approaches have addressed source reliability, factuality of reporting or political bias, by assembling information from multiple external and social media sources, metadata and/or content-based features~\cite{baly-etal-2020-written,baly-etal-2019-multi,baly-etal-2018-predicting,panayotov2022greener,da2023overview,nakov2023overview,bozhanova2021predicting}. 
Unfortunately, methodologies relying on social media metadata can not longer be reproduced at scale given the current access restrictions.

A recent research shifting from the social media and text-based approach, is presented in~\cite{burdisso-etal-2024-reliability}. 
Burdisso \textit{et al.}, proposed a highly performing and robust graph-based methodology to score news media reliability. 
Their method considers the longitudinal interactions on the web to learn a reliability value from their source neighbors.  
Based on the research evidence that neighboring properties can be spread among news media outlets, 
we extend their work and we propose to address the following research question: 
to what extent it is possible to profile news media outlets (i.e., different properties) based solely on their interaction with other media sources? 
To address this question, in this paper we focus on two challenging media bias descriptors: factuality of reporting and political bias. We choose to extend Burdisso \textit{et al.} methodology given that it is both language and content independent (political, religious, racial, etc.), it can be applied at a larger scale and their 17k English news outlets dataset is publicly available.

Our main contributions are as follows: (a) we show that it is possible to predict/estimate bias descriptors, i.e.,  political bias and factual reporting; of the source media based on their interactions with other sources (outperforming the baseline); (b) we validate the robustness of our approach on the publicly available dataset from the CLEF CheckThat! challenge, specifically collected to classify political bias with currently active news outlets, and we established a new SOTA result; (c) we release the biggest dataset at the source media level with standard political bias and factual reporting labels.

\section{Related Work}
The bias in news media is a pervasive and ubiquitous problem~\cite{Hamborg2023,hanimann2023believing,stromback2020news}. 
The need for applied research on news media descriptors has increased since 2000 due to the generalized adoption of social media platforms, and the proliferation of tools that facilitate both websites and news-content creation~\cite{fang2024bias,billard2023designing}. 
Bias in news media has a wide descriptors spectrum~\cite{Hamborg2023,hanimann2023believing,wessel2023introducing}, for example Racial Bias refers to preferences of coverage or not of events related to minorities or group of individuals~\cite{pope2018awareness}.
Gender Bias refers to the inclination towards one gender over another, resulting in unequal treatment, coverage and perception~\cite{van2020gender,asr2021gender}.
Political Bias, refers to partial representation of political issues or tendency to favor a particular political ideology. 

A significantly large NLP community has reported advances on news media bias at the article level (i.e., based on content), also referred as bias at the event-label or a short-term bias on a selected event~\cite{Hamborg2023}.
However, in this paper, we contribute towards the news media source profiling (i.e., at the source level). We focus our research work on the following two long-term bias descriptors:

\noindent
\textbf{Factual Reporting} Recent task challenges, particularly the \textit{CheckThat!} Lab challenge at CLEF 2023, have addressed the factuality of reporting based on three classes (\textit{High}, \textit{Mixed} and \textit{Low}) at the article level~\cite{nakov2023overview}. Submitted models range from traditional supervised models (such as SVMs, Random Forest, gradient Boost) to Deep Learning-based ones~\cite{nakov2023overview,nakov2023overview,Leburu-Dingalo2023UBCS}. 
Due to the challenging nature to perform factuality assessment, graph-based models emerged to address the problem disclosing better performance when combined with text-based approaches~\cite{fairbanks2018credibility,baly-etal-2020-written,baly-etal-2018-predicting,panayotov2022greener}. 
Fairbanks \textit{et al.}, \cite{fairbanks2018credibility} proposed a structural model based on the metadata from the article's news web links. Their findings revealed that credibility, a descriptor in close relation with factual reporting, is harder to determine from merely the content.
Baly \textit{et al.} \cite{baly-etal-2018-predicting} analyzed the factual reporting focusing on the source media. Their approach used text-based features from articles content and metadata including Wikipedia pages, Twitter, URL-related features (domain, orthography, char n-grams), and Web traffic (Alexa service). 
Also targeting the factuality at the media level, Panayotov \textit{et al.},~\cite{panayotov2022greener} proposed to model the factuality of reporting using graph neural network and similarity between news media based on their audience overlap. 
Although the latest models revealed significant improvements at the media level, the methods in \cite{baly-etal-2018-predicting,panayotov2022greener} rely on the Alexa website ranking and web traffic information, which is now discontinued.

More recent approaches are focusing on state-of-the-art LMs and LLMs, from adversarial training, ensemble of models based on RoBERTa or GPTs~\cite{li2023cucplus,tran2023accenture}. Li \textit{et al.}, heuristics on adversarial training revealed the importance of semantics in the title and the summary of the news captured at the beginning and end of the article. Their best performing political inference results from a majority voting from four implemented models from which, two are RoBERTa-based. Tran \textit{et al.}, examined the impact of imbalanced training data between \textit{High}, \textit{Mixed} and \textit{Low} factual reporting. The authors introduced a RoBERTa-based back-translation framework that significantly surpassed the baseline performance. Their approach ranked among the top three performers at the \textit{CheckThat!} Lab challenge in 2023. 
To the best of our knowledge, the state-of-the-art methodology in media profiling, outperforming ensembles of content-based and external data was recently introduced in~\cite{burdisso-etal-2024-reliability}. Burdisso \textit{et al.}, propose an hyperlink-interactions graph to infer News source reliability degree (a continuous value) based on reinforcement learning techniques. In addition to the standing performance, authors contribute with the largest reported dataset in source media profiling with 17k English-speaking news outlets.

\noindent
\textbf{Political Bias} %
In the recent years, the inference of political bias at the outlet level has been approached by applying SVMs, CatBoost and applied oversampling techniques, mostly enhancing content-features from articles \cite{baly-etal-2020-written,da2023overview,azizov2023frank}. 
Baly \textit{et al.}, \cite{baly-etal-2020-written,baly-etal-2018-predicting} proposed a framework based on SVMs reporting significant results when complementing content-based data with Wikipedia and social media metadata. 
Recently, Azizov \textit{et al.}, \cite{azizov2023frank} proposed a majority voting ensemble of CatBoost models and TF-IDF, showing better performance than LM-frameworks at the \textit{CheckThat!} lab challenge at CLEF 2023~\cite{da2023overview} given a benchmark dataset with three political classes (\textit{Left}, \textit{Center}, \textit{Right}).
In Panayotov \textit{et al.}~\cite{panayotov2022greener}, the political bias was modeled using a graph neural network augmented with audience/social media data.
Graph-based approaches showed evidence that metadata capturing information other than the article content improved classification of political stance. 
Given the still open challenge to accurately infer political bias at the news source level, more recent approaches are exploring the pertinence of using LMs~\cite{tran2023accenture,wessel2023introducing}. 
Tran \textit{et al.}~\cite{tran2023accenture}, analyzed and addressed the three-class (\textit{Left}, \textit{Center}, \textit{Right}) imbalance by translating to Spanish and back to English the classes with less articles. Then, they fine-tuned RoBERTa English-large, and performed a majority voting at the article-level to infer the news source political leaning, showing a significant performance above the baseline.
Wessel \textit{et al.}, \cite{wessel2023introducing} proposed a framework using transformers to infer 9 bias descriptors. For the case of political bias, the original bias annotation provided at the outlet level is transformed into two classes bias and not-bias.
Despite the 2 million political news articles used in this work, they were exclusively gathered from the top 11 most popular US media outlets.
Authors concluded that cognitive and political bias at the content-level are the most challenging bias descriptors to detect, in contrast with for example gender or racial bias.

Although some approaches show significant improvement over majority baselines, the robustness and scalability of the models is not sufficient to consider the factual reporting and political bias problem solved. Contrary to previous research that depends on content, audience feedback, and/or metadata, in this paper we extend a very recent work that models the problem in a scalable fashion relaying on network interactions among the news sources~\cite{burdisso-etal-2024-reliability}. Following sections describe the proposed methodology and obtained results.

\section{Methodology and Strategies}

In order to validate our research question and based on the evidence presented by Burdisso \textit{et al.} that longitudinal interactions 
can spread the news media reliability degree among their neighbors~\cite{burdisso-etal-2024-reliability}, we extend their work to address factual reporting and political bias. 

The introduced approach consists of first building a news media graph from the WWW and then applying different reinforcement learning strategies to infer the reliability values. More precisely, constructing a weighted directed graph $G=\langle S, E, w\rangle$  where there is an edge $(s, s') \in E$ if source $s$ contains articles (hyper) linked to $s'$ and where the weight $w(s, s') \in [0, 1]$ is the proportion of total hyperlinks in $s$ linked to $s'$.\footnote{Note that this simple hyperlink-based representation is also implicitly capturing content-based references to and from other sources.} In this work, we hypothesize that the political bias and factual reporting of sources $s$ can be estimated from the sources it interacts with, by inheriting their properties.
%(e.g. sources interacting primarily with extreme-\textit{Left} sources have higher chances of being extreme-\textit{Left} too, see Fig.~\ref{fig:example}).

\

Following the original work in~\cite{burdisso-etal-2024-reliability}, we model the estimation as a Markov Decision Process (MDP) $\langle\mathbb{S}, A, P, r\rangle$ such that:
(1) The set of states $\mathbb{S}$ are all the news outlets websites ---\textit{i.e.} $\mathbb{S} = S$;
(2) The set of actions $A$ contains only one element, the \textit{"move to a different news media website"} action;
(3) The probability $P$ of moving from the origin $s$ to $s'$ will be given by the proportion of hyperlinks in $s$ connecting to $s'$ ---\textit{i.e.} we have $P(s, s') = w(s, s')$; and
(4) The reward $r$ of moving to another news source ($s'$) is determined only by the origin source($s$), and it will be positive or negative depending on the known property ---\textit{e.g} $r(s) = 1$ if we know for this $s$ we have \textit{Right} or \textit{High}, for political bias or factual reporting, respectively; $r(s) = -1$ if $s$ is \textit{Left} or \textit{Low}, for political bias or factual reporting, respectively; $r(s) = 0$ otherwise.
%5-fold cross-validation
Finally, the property (political bias or factual reporting level) value for all news sources $s$ in the graph will be estimated by a function $\rho(s)$ following 4 different strategies:
%, formulated as follows:

\begin{itemize}
    \item \textbf{F}-\textit{property}: The property value is proportional to the \textit{expected} perceived reward given by the following Bellman equation where $\pi$ is  the unique policy (i.e. the probability of taking action $a \in A$ in state $s$) and $\gamma \in [0,1)$ the discount factor:\footnote{The discount factor controls the distance of looking back/forward; $\gamma\approx0$ focuses mostly on present reward $r(s)$, while $\gamma\approx1$ considers all history/future to compute $\rho(s)$.} 
   % (with $V(s) = \rho(s)$):
    \begin{equation}
    \label{eq:value}
	\rho(s) = \sum_{s'\in\mathbb{S}}{P^\pi(s, s')[r(s') + \gamma \rho(s')]}
    \end{equation}
    That is, under this strategy, the value of source $s$ will be inherent from the sources it connects in the \textbf{F}uture.

    \item \textbf{P}-\textit{property}: The property value is interpreted a proportion of the accumulated perceived reward, i.e., the value is inherited by the sources that lead to it in the \textbf{P}ast. The value is thus, giving by the following the reverse Bellman equation: 
    \begin{equation}
    \label{eq:reverse}
    \rho(s) = r(s) + \gamma\sum_{s'\in\mathbb{S}}{P^\pi(s', s)\rho(s')}
    \end{equation}

    \item \textbf{FP}-\textit{property}: This strategy strategy combines the previous two strategies by considering \textbf{F}uture and \textbf{P}ast 
    information. A source $s$ increases its positive value $\rho(s)$ as more positive sources link to it ($\rho_\textbf{P}^+(s)$), while losing value as it links to more negative sources ($\rho_\textbf{F}^-(s)$).\footnote{Algorithm 1 and Algorithm 2 in \cite{burdisso-etal-2024-reliability} detail how these updates are applied.} Thus, $\rho(s)$ is simply defined as:

    \begin{equation}
    \label{eq:fp}
	\rho(s) = \rho_\textbf{F}^-(s) + \rho_\textbf{P}^+(s)
    \end{equation}
    
    \item \textbf{I}-\textit{property}: \textbf{I}nvestment Strategy (invest and collect credits) consisting of two iterative steps, repeated $n$ times: (1) all  sources invest their property value to the neighboring sources proportionally to the strength of their links ($w(s, s')$) following Equation~\ref{eq:totalcredits}, (2) sources collect the credits back proportionally to the investment and update its own property value following Equation~\ref{eq:investment}.
    
    \begin{equation}
    totalcredits(s) = \sum_{s'\in S}{w(s', s)\cdot\rho(s')}
    \label{eq:totalcredits}
    \end{equation}

    \begin{equation}
    \rho(s) = \rho(s) + \sum_{s'\in S}{w(s, s')\cdot credits_{s}(s')}
    \label{eq:investment}
    \end{equation}

    where credits are distributed among investors $s'$, in proportion to their contribution to $s$, i.e, $credits_{s'}(s) = w_{s'}(s) \cdot totalcredits(s)$.

\end{itemize}

\subsection{Datasets}

There are several attempts to unify existing datasets to assess  Bias in news media. Recently, a unified bias dataset was presented including several Bias descriptors~\cite{wessel2023introducing}, nevertheless, the collection of articles, sentences, comments, etc., are on one hand targeting rather short-term bias (text-based), and on the other hand large part of the data do not have URLs to existing news media sources. 
Recently, \cite{burdisso-etal-2024-reliability} released the largest dataset with URLs annotated with reliability labels constructed by collecting and consolidating annotations from different sources.
In this work, we follow a similar process as described by the authors in \cite{burdisso-etal-2024-reliability} to build our own dataset with political bias and factual reporting annotation, which we refer to as ``MBFC''. 

\

\textbf{MBFC}. Following the methodology described in~\cite{burdisso-etal-2024-reliability}, we crawled 3920 news media URL domains from the \textit{Media Bias/Fact Check} (MBFC)\footref{note1} 
service including annotated bias descriptors that are further transformed and normalized into political and factual reporting labels as follows: for Political bias, the final normalized categories are \textit{Left, Center, Right}; for the case of Factual Reporting, labels include \textit{High} (which aggregated high and very high), \textit{Mixed} and \textit{Low} (which aggregates low and very low).

\

\textbf{CLEF \textit{CheckThat!}}. Additionally, in order to compare results with previously published approaches, we use the dataset released for the CLEF 2023 \textit{CheckThat!} lab which focused on political bias identification. This dataset contains a total of 1023 news media URL domains with political bias labels crawled from \textit{allsides}\footref{noteallsides}, a website that gathers news articles with balanced representation of the different political perspectives. The data is officially divided into fixed train, dev, and test set splits containing 817 \textit{(Left-216, Center-296 and Right-305)}, 104 \textit{(Left-31, Center-34 and Right-39)}, and 102 \textit{(Left-25, Center-29 and Right-48)} news sources, respectively.
More details about the data and the labeling process can be found in \cite{da2023overview}. Table \ref{tab:dataset-mbfc} summarizes the label distribution and size of both introduced datasets. 

\begin{table}[!t]
\caption{Label distribution on both datasets.}\label{tab:dataset-mbfc}
\small
\centering
\begin{tabular}{l|c@{~~}c@{~~}c@{~~}|@{~~}c@{~~}c@{~~}c@{~~}|@{~~}c@{~~}}
\toprule
\textbf{Dataset} & \multicolumn{3}{c|@{~~}}{\textbf{Political Bias}} &\multicolumn{3}{@{~~}c@{~~}|@{~~}}{\textbf{Factual Rep.}} & \\
 & \textit{Left} & \textit{Center} & \textit{Right} & \textit{Low} & \textit{Mixed} & \textit{High} & \textbf{Total} \\
\midrule

MBFC (ours) & 2078 & 763 & 1079 & 408 & 1391 & 2121 & 3920 \\
CLEF \textit{CheckThat!} & 272 & 359 & 392 & - & - & - & 1023 \\
\bottomrule
\end{tabular}
\end{table}

\section{Experiments and Results}

In this work we used the graph $G$ built in~\cite{burdisso-etal-2024-reliability} consisting of 17K news sources obtained after processing 100M news articles from Common Crawl News.
Following ~\cite{baly-etal-2018-predicting,burdisso-etal-2024-reliability} we report 5-fold cross-validation evaluation results on our MBFC datasets, whereas for CLEF’s \textit{CheckThat!} we report results on the official test set.
In order to estimate the factual score of reporting from the graph, we first convert the factuality/bias ground truth labels from the training set into rewards as follows: $r(s) = 1$ if the media label is \textit{High/Right}, $r(s) = -1$ if \textit{Low/Left}, and $r(s) = 0$ otherwise.
Then, at inference time, sources $s$ are classified with the label \textit{Right/High} if $\rho(s) > 0$ and \textit{Left/Low} otherwise.
Even though one limitation of the proposed strategies is that they are essentially binaries, in order to compare results in \textit{CheckThat!} three-label classification task, we use the official dev set to find an $\epsilon$ value to classify sources $s$ as follow: \textit{Left/Low} if $\rho(s) < -\epsilon$; \textit{Right/High} if $\rho(s) > \epsilon$; \textit{Center/Mixed} otherwise.
More precisely, we selected the hyper-parameters $\epsilon=3\mathrm{e}{-3}$, $\gamma=0.15$ (Equation~\ref{eq:value}, \ref{eq:reverse}, \ref{eq:fp}), and $n=2$ (Equation~\ref{eq:investment}) after performing a grid search maximizing the Macro avg. F1 score with $\epsilon \in [1\mathrm{e}{-3}, 1\mathrm{e}{-1}]$ ($1\mathrm{e}{-3}$ increments), $\gamma\in [0.05, 0.95]$ ($0.05$ increments), $n \in [1, 10]$, respectively.

\begin{table}[t] \caption{5-fold cross-validation average results for Political Bias and Factual Reporting classification. 
    The best-performing values are \underline{\textbf{underlined}}, while the 2nd-best results appear in \textbf{bold} font.
    }
    \label{tab:results-cv}
    \centering
    \small
    \begin{tabular}{c@{~~}c@{~~~}c@{~~~}c@{~~~}c@{~~~}c}
        \toprule
        &\multirow{2}{*}{\textbf{Strategy}} & \multicolumn{3}{c}{\textbf{F$_1$ score}} &  \\
        \cmidrule(lr){3-5}
        \textbf{Task} &  & \textit{Macro avg.} & \textit{High/Right} & \textit{Low/Left} & \textbf{Accuracy} \\
        \midrule

        \multirow{7}{*}{\rotatebox{90}{\textbf{Factual Rep.}}} & \textit{Majority} & 38.94 $\pm$ 0.04 & 87.88 $\pm$ 0.09 & 0.00 $\pm$ 0.00 & 83.84 $\pm$ 0.16 \\
        & \textit{Random} & 36.44 $\pm$ 0.88 & 65.69 $\pm$ 1.64 & 7.18 $\pm$ 1.45 & 49.93 $\pm$ 1.69 \\
        \cmidrule(lr){2-6}
        & \textbf{F-}Factuality & 57.60 $\pm$ 4.38 & 95.00 $\pm$ 0.97 & 20.19 $\pm$ 7.86 & 90.60 $\pm$ 1.76 \\
        & \textbf{P-}Factuality & \textbf{85.13} $\pm$ 2.73 & \textbf{98.70} $\pm$ 0.35 & \textbf{71.55} $\pm$ 5.15 & \textbf{97.52} $\pm$ 0.66 \\
        & \textbf{F\textbf{P-}}Factuality & 71.35 $\pm$ 2.33 & 96.76 $\pm$ 0.65 & 45.93 $\pm$ 4.09 & 93.89 $\pm$ 1.19 \\
        & \textbf{I-}Factuality & \underline{\textbf{87.99}} $\pm$ 4.60 & \underline{\textbf{99.02}} $\pm$ 0.43 & \underline{\textbf{76.96}} $\pm$ 8.79 & \underline{\textbf{98.12}} $\pm$ 0.81 \\

        \midrule
        \midrule
        
        \multirow{7}{*}{\rotatebox{90}{\textbf{Political Bias}}} & \textit{Majority} & 38.04 $\pm$ 0.07 & 0.00 $\pm$ 0.00 & 76.08 $\pm$ 0.14 & 65.40 $\pm$ 0.18 \\
        & \textit{Random} & 45.42 $\pm$ 1.84 & 30.29 $\pm$ 2.86 & 60.55 $\pm$ 1.75 & 49.65 $\pm$ 1.73 \\
        \cmidrule(lr){2-6}
        & \textbf{F-}Political & 60.42 $\pm$ 3.74 & 41.56 $\pm$ 6.27 & 79.29 $\pm$ 1.42 & 69.44 $\pm$ 2.30 \\
        & \textbf{P-}Political & \textbf{74.08} $\pm$ 2.31 & \textbf{65.80} $\pm$ 3.23 & \textbf{82.36} $\pm$ 1.39 & \textbf{76.73} $\pm$ 1.95 \\
        & \textbf{F\textbf{P-}}Political & 64.90 $\pm$ 3.15 & 52.47 $\pm$ 4.82 & 77.33 $\pm$ 1.94 & 69.34 $\pm$ 2.55 \\
        & \textbf{I-}Political & \underline{\textbf{77.77}} $\pm$ 2.45 & \underline{\textbf{70.97}} $\pm$ 3.39 & \underline{\textbf{84.56}} $\pm$ 1.54 & \underline{\textbf{79.85}} $\pm$ 2.12 \\

        % ExpsetB

        \bottomrule
    \end{tabular}
   
\end{table}

\subsection{Factuality of Reporting}

Table~\ref{tab:results-cv} shows the results from the 5-fold cross-validation for Factual Reporting. The baseline for comparison includes Random and Majority class classification. 
The \textbf{F-}Factuality strategy performed at 57.60 F1-score overall, for the individual classes \textit{Low} Factual reporting performance is 95.00 F1-score and 20.19 for \textit{High} Factual Reporting. For all cases there is significant improvement with respect to the baselines.
For \textbf{P-}Factuality F1-score performance is 85.13, and 98.7 and 71.55 for the \textit{Low} and \textit{High} classes. The significantly high performance reveals that indeed the graph with past reward strategy captures close interacting networks on both sides, \textit{High} score and \textit{Low} score of factual reporting. 
The strategy \textbf{F\textbf{P-}}Factuality performs at 71.35 F1-score, although it outperforms \textbf{F-}Factuality and the baselines, it remains behind \textbf{P-}Factuality.
Finally, the \textbf{I-}Factuality strategy outperforms all the other strategies up to 87.99 F1-score, 76.96 for class High and 99.02 for the class Low. 
The results show that for the case of \textbf{I-}Factuality (the invest and collect strategy), the gathered information from the hyperlinks and its neighbors can accurately capture the level of factuality, significantly better for the class Low.

\textbf{I-}Factuality accurately identifies almost all sources with \textit{Low} Factual Reporting, which is indeed a key contribution of this paper. We assume that high performance of the reward value might be due to capturing unintentionally the lifespan of a news media domain, which has been reported as a high contributor in the identification of disinformative websites~\cite{hounsel2020identifying}. Both strategies \textbf{P-}Factuality and \textbf{I-}factuality are highly performing on F1-score and Accuracy, similarly to findings on Reliabilty of news media in~\cite{burdisso-etal-2024-reliability}, disclosing an accurate profiling of Bias given only their network interactions overtime.

\subsection{Political Bias}

\begin{table}[!t]
\caption{Results on CLEF's \textit{CheckThat!} dataset on Political Bias of news media. MAE: Mean Absolute Error. The smaller MAE value translates into better predictions.}
\label{tab:results-clef}
\centering
\begin{tabular}{lc@{~~}c@{~~}c@{~~}}
\toprule
\textbf{Team} &  \textbf{MAE}($\downarrow$) & \textbf{F$_1$ score}($\uparrow$) & \textbf{Accuracy}($\uparrow$) \\
\midrule
Baseline~\cite{da2023overview} & 0.902 & - & - \\
Awakened & 0.765 & - & - \\
Accenture~\cite{tran2023accenture} &  0.549 & 0.625 & 0.627 \\
Frank~\cite{azizov2023frank} & 0.320 & 0.727 & 0.725 \\ % 0.62  0.621 in original stats, then voting gives the results on the table
\midrule
%95% confidence level, significance with respect to Top performance Frank 
\textbf{F-}Political & 0.333 & 0.632 & 0.667 \\ % Not significant
\textbf{P-}Political & \textbf{0.238} & \textbf{0.760} & \textbf{0.762} \\ % Significant
\textbf{F\textbf{P-}}Political & 0.309 & 0.670 & 0.690 \\ % Not significant
\textbf{I-}Political & \underline{\textbf{0.214}} & \underline{\textbf{0.784}} & \underline{\textbf{0.786}} \\ %Significant
\bottomrule
\end{tabular}
\end{table}

\begin{figure*}[t]
    \centering
    \includegraphics[width=.7\linewidth]{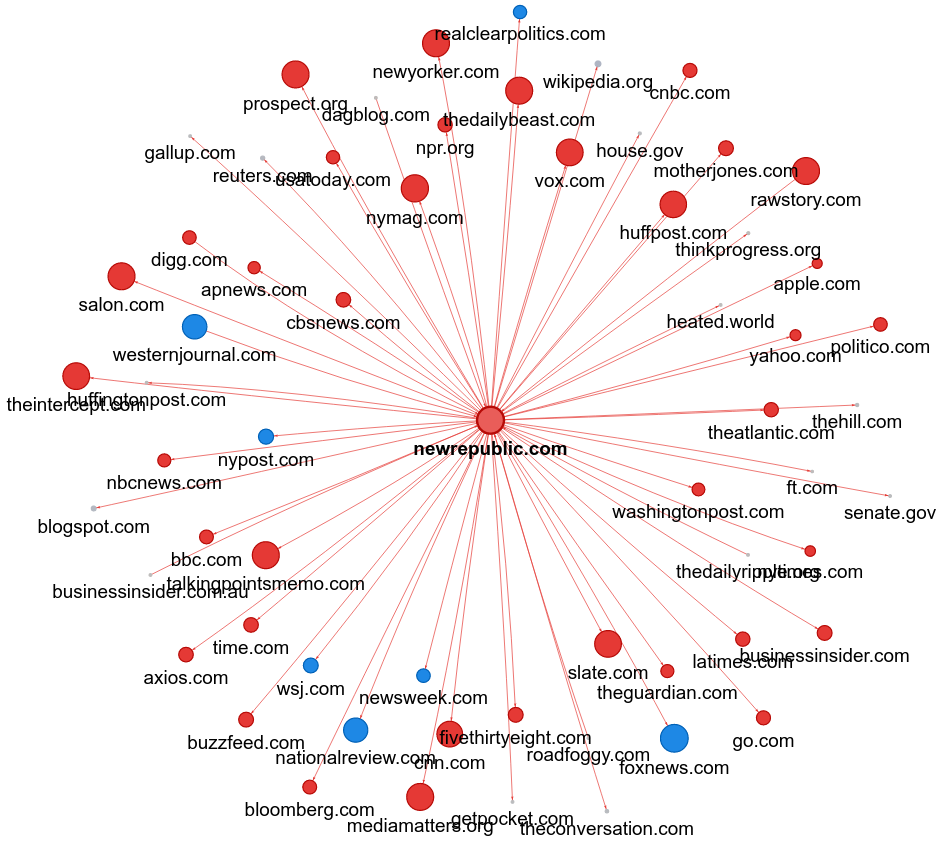}
    \caption{Example showing how \texttt{newrepublic.com} relates with neighboring news sources. \textit{Left} and \textit{Right} wing sources are colored red and blue respectively, in addition, size of the node reflects the degree of the bias (learned by our \textbf{I}-political strategy). We can see that since \texttt{newrepublic.com} interacts mostly with \textit{Left}-wing sources, its final bias degree ended up being considerable \textit{Left}-wing.}
    \label{fig:example}
\end{figure*}

\noindent
\textbf{Results on MBFC}.
Table \ref{tab:results-cv} shows the 5-fold cross-validation results for F1-score and Accuracy, we included two baselines Random, and Majority class for comparison. 
For the political leaning the \textbf{F-}Political performs at 60.42 F1-score, and 79.29 F1-score for Right at the class level, showing a modest improvement over the baseline (76.08). For \textbf{P-}Political the overall F1-score performance is 74.08, with 65.8 for the class \textit{Left} and 82.36 for the class \textit{Right}. For the combined \textbf{F\textbf{P-}}Political the F1-score of 64.90 outperforms the \textbf{F-}Political but does not improve the \textbf{P-}Political performance, for both the overall and the class level, which indicates that past information contributes more to the predictions.
The best performing strategy is \textbf{I-}Political performing at 77.77 F1-score and, 70.97 and 84.56 for the classes \textit{Left} and \textit{Right} respectively. At the class level, our results on political bias show significantly better performance on the class \textit{Right}. 
Figure~\ref{fig:example} shows part of the graph for the news media source \texttt{www.newrepublic.com}, where the values are estimated with \textbf{I-}political. 
The size of the node is proportional to their political bias, as newrepublic predominantly engages with \textit{Left}-wing sources, its final value leaned significantly towards the \textit{Left} (red).

\

\noindent
\textbf{Results on the CLEF \textit{CheckThat!}}.
Table~\ref{tab:results-clef} shows the F1-score performance and the official scoring metric MAE (Mean Absolute Error) for the Labs at CLEF 2023. The political labels were coded as ordinal values (\textit{Left}-0, \textit{Center}-1, \textit{Right}-2), a smaller MAE value translates into better predictions from the proposed models. The baseline with MAE of 0.902 uses an SVM classification model based on N-Grams. The top performed participating model \cite{azizov2023frank} achieved a MAE of 0.320, outperforming the baseline and the other participating models. However, our proposed strategies (\textbf{P}-Political and \textbf{I}-Political) outperform the best-performing participating model in all reported metrics the top (MAE, F1-score and Accuracy). 
The MAE top performance (smaller MAE) indicates that the miss-predictions are less severe (from \textit{Center} to the extremes or vice-versa), otherwise inferences will result on a higher penalization if predicting completely opposite extremes \textit{Left} $\leftrightarrow$ \textit{Right}.

\section{Conclusions}

This research extends the methodology proposed in~\cite{burdisso-etal-2024-reliability} by addressing long-term news media profiling, contrasting with approaches focused solely on short-term bias. Our experiments on two challenging bias descriptors—factual reporting and political bias—utilize four reinforcement learning strategies for classification performance evaluation.
We provide compelling evidence supporting the longitudinal view of news media and their web interactions as a robust and scalable proxy for profiling, particularly regarding political bias and factual reporting. Concretely, performed experiments show that the proposed approach allows superior performance in estimating outlet media bias descriptors compared to baseline methods. Furthermore, we present promising results from comparisons with other participating models submitted to the CLEF 2023 \textit{CheckThat!} lab, designed for inferring political bias in currently active news outlets. Our approach surpasses top results in both F1-score and the official MAE performance measure, establishing a new SOTA result for this particular task. Finally, as an additional contribution, we release the largest dataset at the source media level, annotated with standard political bias and factual reporting labels.

As part of future efforts, we aim to investigate the dynamics of political bias changes over time within news media, such as shifts from center to extreme positions. Additionally, we plan to explore the integration of other bias descriptors, such as press freedom, in multi-task bias identification.

\begin{credits}
\subsubsection{\ackname} This work was supported by \href{https://www.project-criteria.eu}{CRiTERIA}, EU project funded under the Horizon 2020 program, grant agreement number 101021866.

\subsubsection{\discintname}
%It is now necessary to declare any competing interests or to specifically
%state that 
The authors have no competing interests. 
%(optional) acknowledgments\footnote{If EquinOCS, our proceedings submission
%system, is used, then the disclaimer can be provided directly in the system.},
%for example: The authors have no competing interests to declare that are
%relevant to the content of this article. Or: Author A has received research
%grants from Company W. Author B has received a speaker honorarium from
%Company X and owns stock in Company Y. Author C is a member of committee Z.
\end{credits}
%
% ---- Bibliography ----
%
% BibTeX users should specify bibliography style 'splncs04'.
% References will then be sorted and formatted in the correct style.
%
\bibliographystyle{splncs04}
\bibliography{main}

\begin{thebibliography}{10}
\providecommand{\url}[1]{\texttt{#1}}
\providecommand{\urlprefix}{URL }
\providecommand{\doi}[1]{https://doi.org/#1}

\bibitem{asr2021gender}
Asr, F.T., Mazraeh, M., Lopes, A., Gautam, V., Gonzales, J., Rao, P., Taboada, M.: The gender gap tracker: Using natural language processing to measure gender bias in media. PloS one  \textbf{16}(1),  e0245533 (2021)

\bibitem{azizov2023frank}
Azizov, D., Liang, S., Nakov, P.: Frank at checkthat! 2023: Detecting the political bias of news articles and news media. Working Notes of CLEF  (2023)

\bibitem{baly-etal-2018-predicting}
Baly, R., Karadzhov, G., Alexandrov, D., Glass, J., Nakov, P.: Predicting factuality of reporting and bias of news media sources. In: Proceedings of the Conference on Empirical Methods in Natural Language Processing. pp. 3528--3539 (oct 2018)

\bibitem{baly-etal-2020-written}
Baly, R., Karadzhov, G., An, J., Kwak, H., Dinkov, Y., Ali, A., Glass, J., Nakov, P.: What was written vs. who read it: News media profiling using text analysis and social media context. In: Proceedings of the Association for Computational Linguistics (2020). \doi{10.18653/v1/2020.acl-main.308}

\bibitem{baly-etal-2019-multi}
Baly, R., Karadzhov, G., Saleh, A., Glass, J., Nakov, P.: Multi-task ordinal regression for jointly predicting the trustworthiness and the leading political ideology of news media. In: Proceedings of the North {A}merican Chapter of the Association for Computational Linguistics (2019). \doi{10.18653/v1/N19-1216}

\bibitem{billard2023designing}
Billard, T.J., Moran, R.E.: Designing trust: Design style, political ideology, and trust in “fake” news websites. Digital Journalism  \textbf{11}(3),  519--546 (2023)

\bibitem{bozhanova2021predicting}
Bozhanova, K., Dinkov, Y., Koychev, I., Castaldo, M., Venturini, T., Nakov, P.: Predicting the factuality of reporting of news media using observations about user attention in their youtube channels. In: Proceedings of the International Conference on Recent Advances in Natural Language Processing. pp. 182--189 (2021)

\bibitem{burdisso-etal-2024-reliability}
Burdisso, S., Sanchez-Cortes, D., Villatoro-Tello, E., Motlicek, P.: Reliability estimation of news media sources: Birds of a feather flock together. In: Proceedings of the North American Chapter of the Association for Computational Linguistics (2024), \url{https://aclanthology.org/2024.naacl-long.383}

\bibitem{da2023overview}
Da~San~Martino, G., Alam, F., Hasanain, M., Nandi, R.N., Azizov, D., Nakov, P.: Overview of the clef-2023 checkthat! lab task 3 on political bias of news articles and news media. Working Notes of CLEF  (2023)

\bibitem{esteves2018belittling}
Esteves, D., Reddy, A.J., Chawla, P., Lehmann, J.: Belittling the source: Trustworthiness indicators to obfuscate fake news on the web. In: Proceedings of the First Workshop on Fact Extraction and {VER}ification ({FEVER}). pp. 50--59 (2018)

\bibitem{fairbanks2018credibility}
Fairbanks, J., Fitch, N., Knauf, N., Briscoe, E.: Credibility assessment in the news: do we need to read. In: Proc. of the MIS2 Workshop held in conjuction with 11th Int’l Conf. on Web Search and Data Mining. pp. 799--800. ACM (2018)

\bibitem{fang2024bias}
Fang, X., Che, S., Mao, M., Zhang, H., Zhao, M., Zhao, X.: Bias of ai-generated content: an examination of news produced by large language models. Scientific Reports  \textbf{14}(1),  1--20 (2024)

\bibitem{gezici2022quantifying}
Gezici, G.: Quantifying political bias in news articles (2022)

\bibitem{Hamborg2023}
Hamborg, F.: Media Bias Analysis, pp. 11--53. Springer Nature Switzerland, Cham (2023). \doi{10.1007/978-3-031-17693-7_2}

\bibitem{hanimann2023believing}
Hanimann, A., Heimann, A., Hellmueller, L., Trilling, D.: Believing in credibility measures: reviewing credibility measures in media research from 1951 to 2018. International journal of communication  \textbf{17},  214--235 (2023)

\bibitem{hounsel2020identifying}
Hounsel, A., Holland, J., Kaiser, B., Borgolte, K., Feamster, N., Mayer, J.: Identifying disinformation websites using infrastructure features. In: USENIX Workshop on Free and Open Communications on the Internet (FOCI) (2020)

\bibitem{kamalloo2023evaluating}
Kamalloo, E., Dziri, N., Clarke, C.L., Rafiei, D.: Evaluating open-domain question answering in the era of large language models. In: Proceedings of the North {A}merican Chapter of the Association for Computational Linguistics (2023). \doi{10.18653/v1/2023.acl-long.307}

\bibitem{kulshrestha2019search}
Kulshrestha, J., Eslami, M., Messias, J., Zafar, M.B., Ghosh, S., Gummadi, K.P., Karahalios, K.: Search bias quantification: investigating political bias in social media and web search. Information Retrieval Journal  \textbf{22},  188--227 (2019)

\bibitem{Leburu-Dingalo2023UBCS}
Leburu-Dingalo, T., Thuma, E., Motlogelwa, N., Mudongo, M., Mosweunyane, G.: Ubcs at checkthat! 2023: Stylometric features in detecting factuality of reporting of news media. Working Notes of CLEF  (2023)

\bibitem{li2023cucplus}
Li, C., Xue, R., Lin, C., Fan, W., Han, X.: Cucplus at checkthat! 2023: text combination and regularized adversarial training for news media factuality evaluation. Working Notes of CLEF  (2023)

\bibitem{nakov2023overview}
Nakov, P., Alam, F., Da~San~Martino, G., Hasanain, M., Nandi, R., Azizov, D., Panayotov, P.: Overview of the clef-2023 checkthat! lab task 4 on factuality of reporting of news media. Working Notes of CLEF  (2023)

\bibitem{panayotov2022greener}
Panayotov, P., Shukla, U., Sencar, H.T., Nabeel, M., Nakov, P.: Greener: Graph neural networks for news media profiling. In: Proceedings of the 2022 Conference on Empirical Methods in Natural Language Processing. pp. 7470--7480 (2022)

\bibitem{van2020gender}
Van~der Pas, D.J., Aaldering, L.: Gender differences in political media coverage: A meta-analysis. Journal of Communication  \textbf{70}(1),  114--143 (2020)

\bibitem{pope2018awareness}
Pope, D.G., Price, J., Wolfers, J.: Awareness reduces racial bias. Management Science  \textbf{64}(11),  4988--4995 (2018)

\bibitem{stromback2020news}
Str{\"o}mb{\"a}ck, J., Tsfati, Y., Boomgaarden, H., Damstra, A., Lindgren, E., Vliegenthart, R., Lindholm, T.: News media trust and its impact on media use: Toward a framework for future research. Annals of the International Communication Association  \textbf{44}(2),  139--156 (2020). \doi{0.1080/23808985.2020.1755338}

\bibitem{tran2023accenture}
Tran, S., Rodrigues, P., Strauss, B., Williams, E.: Accenture at checkthat! 2023: Learning to detect factuality levels of news sources. Working Notes of CLEF  (2023)

\bibitem{wessel2023introducing}
Wessel, M., Horych, T., Ruas, T., Aizawa, A., Gipp, B., Spinde, T.: Introducing mbib-the first media bias identification benchmark task and dataset collection. In: Proceedings of the International ACM SIGIR Conference on Research and Development in Information Retrieval. pp. 2765--2774 (2023)

\end{thebibliography}

\end{document}